\documentclass[sigconf]{acmart}

\usepackage{multirow} 
\usepackage{graphicx}
\usepackage{url}
\usepackage{float}
\usepackage{subfigure}
\usepackage{bm}
\usepackage{amsmath}
\usepackage{amsfonts}
\usepackage{nicefrac}     
\usepackage{microtype}      
\usepackage{lipsum}
\usepackage[ruled,vlined]{algorithm2e}
\usepackage{color}
\usepackage{booktabs}
\usepackage{epsfig}
\usepackage{epstopdf}

\newcommand{\name}{PPro}

\AtBeginDocument{%
  \providecommand\BibTeX{{%
    \normalfont B\kern-0.5em{\scshape i\kern-0.25em b}\kern-0.8em\TeX}}}

\setcopyright{acmcopyright}
\copyrightyear{2018}
\acmYear{2018}
\acmDOI{XXXXXXX.XXXXXXX}

\acmConference[Conference acronym 'XX]{Make sure to enter the correct
  conference title from your rights confirmation emai}{June 03--05,
  2018}{Woodstock, NY}
\acmPrice{15.00}
\acmISBN{978-1-4503-XXXX-X/18/06}

\begin{document}

\title{Prioritized Propagation in Graph Neural Networks}

\author{Yao Cheng}
\affiliation{%
  \institution{East China Normal University}
  \city{Shanghai}
  \country{China}}
\email{yaocheng_623@stu.ecnu.edu.cn}

\author{Minjie Chen}
\affiliation{%
  \institution{East China Normal University}
  \city{Shanghai}
  \country{China}}

\author{Xiang Li}
\affiliation{%
  \institution{East China Normal University}
  \city{Shanghai}
  \country{China}}

\author{Caihua Shan}
\affiliation{%
  \institution{Microsoft Research Asia}
  \city{Shanghai}
  \country{China}}

\author{Ming Gao}
\affiliation{%
  \institution{East China Normal University}
  \city{Shanghai}
  \country{China}}
  


  

\renewcommand{\shortauthors}{Trovato and Tobin, et al.}

\begin{abstract}
 Graph neural networks (GNNs) have
recently received significant attention.
Learning node-wise message propagation in GNNs aims to set personalized propagation steps
for different nodes in the graph.
Despite the success,
existing methods ignore node priority that can be reflected by node influence and heterophily.
In this paper, we propose a versatile framework \name,
which can be integrated with most existing GNN models and aim to learn prioritized node-wise message propagation
in GNNs.
Specifically, 
the framework consists of three components: a backbone GNN model, a propagation controller to determine the optimal propagation steps for nodes, 
and a weight controller to compute the priority scores for nodes.
We design a mutually enhanced mechanism to compute node priority, optimal propagation step and label prediction. 
We also propose an alternative optimization strategy to learn the parameters in the backbone GNN model and two parametric controllers.
We conduct extensive experiments to compare our framework with other 11 state-of-the-art competitors on 8 benchmark datasets.
Experimental results show that our framework can lead to superior performance in terms of propagation strategies and node representations.
\end{abstract}





\settopmatter{printfolios=true}

\maketitle

\section{Introduction}
\label{introduction}
Graphs are ubiquitous in the real world, such as 
social networks \cite{leskovec2010predicting}, biomolecular structures \cite{zitnik2017predicting} and knowledge graphs~\cite{DBLP:abs-2003-02320}.
In graphs, 
nodes represent entities and edges capture the relations between entities. 
To further enrich information in graphs, 
nodes are usually associated with feature vectors.
For example,
in \emph{Facebook},
a user can connect to many other users, where links represent the \emph{friendship} relation; a user can also have \emph{age}, \emph{gender}, \emph{occupation} as descriptive features. 
Both node features and graph structure provide information sources for graph-structured learning. 
Recently,
graph neural networks (GNNs) \cite{kipf2016semi,hamilton2017inductive,klicpera2018predict} have been proposed,
which can seamlessly integrate the two sources of information and have shown superior performance in a variety of downstream web-related tasks, 
such as web recommendation~\cite{DBLP:abs-1806-01973}, social network analysis~\cite{li2019encoding} and anomaly detection on webs~\cite{liu2021anomaly}.

\begin{figure}[h]
  \centering  \includegraphics[width=\linewidth]{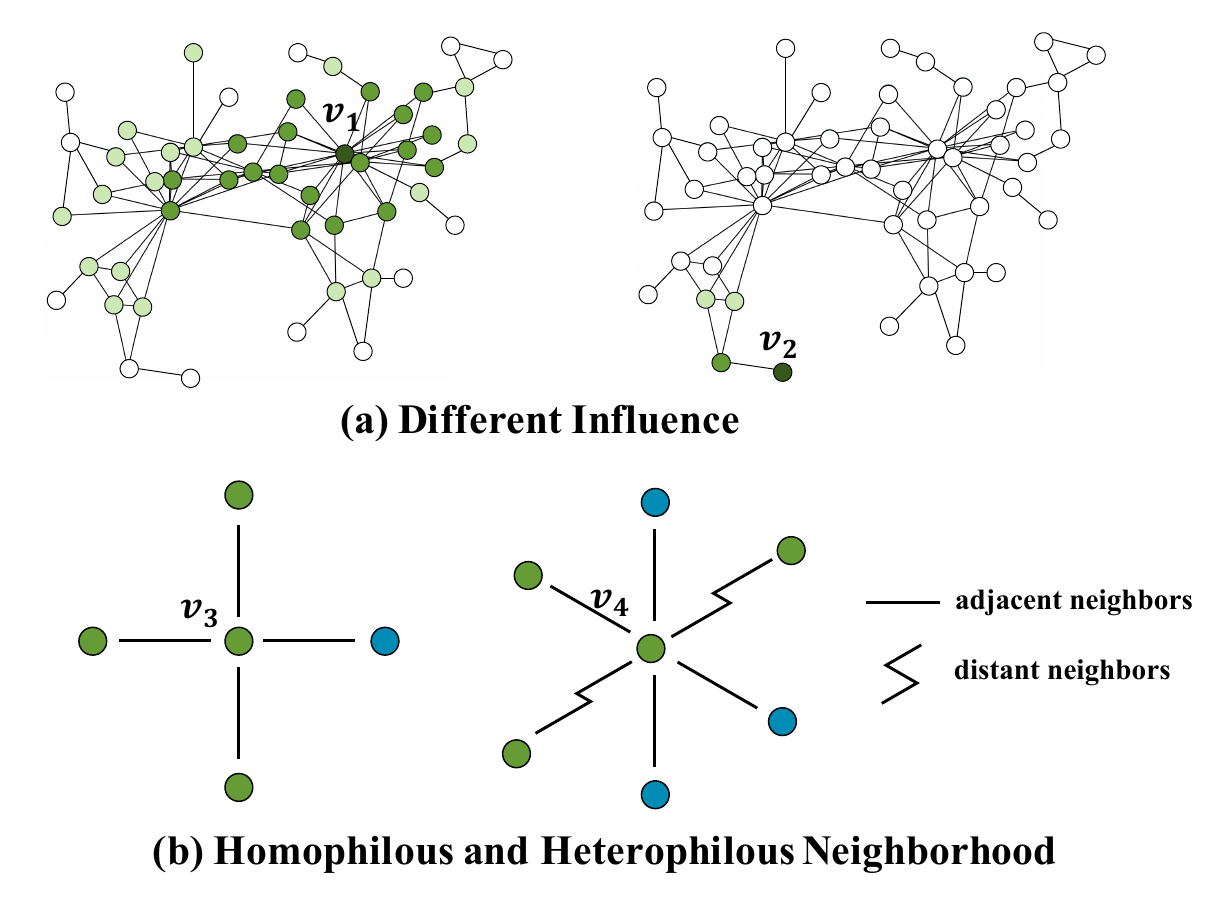}
  \caption{An illustration on node priority}
  \label{fig:example}
\end{figure}
In GNNs, 
the embedding of a node is learned by
aggregating (propagating)
messages from (to) its neighbors. 
Given a pre-set GNN layer $L$,
for each node, 
it can perceive 
messages from neighbors that are 
$L$-hop away. 
With the increase of $L$,
the receptive field of a node gets larger and more information can be aggregated from neighbors to generate the node's embedding.
However,
due to the effect of low-pass convolutional filters used in GNNs,
the embeddings of nodes in the same component of a graph tend to be indistinguishable with more layers stacked.
This is the notorious 
\emph{over-smoothing}~\cite{DropEdge} problem in GNNs.
To tackle the problem,
existing methods utilize various techniques, 
such as 
residual connection~\cite{kipf2016semi}, normalization schemes~\cite{zhao2019pairnorm} and boosting strategy~\cite{sun2020adagcn}.
However,
these methods treat all the nodes in the graph equally. 
Practically, 
nodes in the graph have various local structures. 
For example, 
in Facebook, 
some users play the role of \emph{hub}, 
which link to many other users and are located at the center of a graph; 
some users are \emph{leaf} nodes that have very few connections.
Intuitively, 
nodes with fewer one-hop neighbors need to 
absorb messages from distant nodes for label prediction~\cite{DBLP:abs-2009-03509}. 
However, 
an excessively large neighborhood range could lead to an over-smoothing problem. Further,
it is infeasible to manually set 
propagation steps
for all the nodes in the graph, 
which motivates the design of personalized node-wise message propagation layers.

Recently,
some methods~\cite{xu2018representation,xiao2021learning,wang2022graph} 
are proposed to
learn node-wise message propagation in GNNs.
However,
most of these works
distinguish node local structures only
but ignore the different \emph{priorities} of nodes.
Specifically,
assuming that a hub node is predicted incorrectly,
it will then adversely 
infect a significant number of neighbors and easily contaminate the whole graph.
Therefore,
for hub nodes with high degrees,
they are intuitively more influential than leaf nodes.
For example, as shown in Figure~\ref{fig:example}(a), 
when we set the propagation step $L=2$,
compared with the leaf node $v_{2}$,
the hub node
$v_{1}$ 
has a much larger receptive field,
whose messages  
will propagate to and further influence more nodes in the graph; 
hence,
we have to give higher priority to learning
its message propagation.
Also,
it has been pointed out in~\cite{chen2020simple} that nodes with higher degrees are more likely to suffer from over-smoothing with the increase of convolutional layers,
which further provides evidence for giving larger priorities to high-degree nodes.

On the other hand, 
the priority of a node can also be reflected by its neighborhood.
For a node with \emph{homophilous}~\cite{newman2002assortative} neighborhood (see $v_{3}$ in Figure~\ref{fig:example}(b)),
it tends to share similar characteristics or have the same label as its neighbors, so it is much easier to be classified with small message propagation steps.
In contrast,
for a node whose neighborhood is \emph{heterophilous}~\cite{newman2002assortative} (see $v_{4}$ in Figure~\ref{fig:example}(b)),
it is more likely to be connected with 
nodes that have dissimilar features or different labels.
Therefore, 
its label prediction needs to collect information from distant multi-hop neighbors 
and thus requires a large message propagation step.
Since nodes with heterophilous neighborhood
are difficult to be classified,
we have to pay more attention to those nodes in the learning process.
To sum up, there arises a crucial question to make GNNs better: {\itshape 
How to learn prioritized node-wise message propagation in GNNs?
}

In this paper,
we 
learn \textbf{P}rioritized node-wise message \textbf{Pro}pagation in GNNs
and propose the framework \name.
Specifically,
\name\ consists of three components:
a backbone GNN model, 
a propagation controller 
and a weight controller, 
which are coupled with each other.
For propagation controller,
it is used to 
determine the optimal propagation steps for nodes.
In particular,
we design two strategies for propagation controller: 
{L2B directly learns which step to break message propagation for a node,
while L2U learns the best embedding of a node in all the pre-set steps and returns the corresponding propagation step.
In both strategies,
the controller functions consider node priority by taking factors that could affect node 
influence and heterophily
as input.
For the weight controller,
it computes the priority scores for nodes,
which takes into account the optimal propagation steps of nodes output by the propagation controller.
After node priority scores are calculated,
we integrate them with the widely used supervised loss function in GNNs.
Note that
our framework is easy-to-implement and 
can be plugged in with most existing GNN models.
Finally,
our main contributions in this paper are summarized as:

\noindent$\bullet$
We present a versatile framework \name\ for learning prioritized node-wise message propagation in GNNs.

\noindent$\bullet$
We design a mutually enhanced mechanism to compute 
node priority,
optimal propagation step
and label prediction.
We also propose an alternative optimization strategy to learn parameters.

\noindent$\bullet$
We conduct extensive experiments to verify the effectiveness of our proposed framework.
In particular,
we compare \name\ with 11 other state-of-the-art methods on 8 benchmark datasets to show its superiority.
We also implement \name\ with various backbone GNN models to show its wide applicability.

\section{Related Work}
In this section,
we summarize the related work on GNNs and learning to propagate in GNNs, respectively.

\subsection{Graph Neural Networks}
Recently, GNNs have received significant attention and there have been many GNN models proposed~\cite{kipf2016semi, wu2019simplifying, velivckovic2017graph, xu2018how}.
Existing methods can be mainly divided into two categories: spectral model and spatial model.
The former decomposes
graph signals via graph Fourier transform and convolves on the
spectral components,
while the latter directly aggregates messages from a node's spatially nearby neighbors. 
For example,
the early model GCN~\cite{kipf2016semi} extends the
convolution operation to graphs 
and 
is in essence a spectral model.
The representative spatial model
GraphSAGE~\cite{hamilton2017inductive} aggregates information from a fixed size neighborhood of a node to generate its embedding. 
Graph attention networks (GATs)~\cite{velivckovic2017graph} is also a spatial model, which employs the attention mechanism to learn the
importance of an object’s neighbors and aggregate information
from these neighbors with the learned weights. 
Despite the success,
GNNs could suffer from the over-smoothing problem, 
where embeddings of nodes in the same component of a graph tend to be indistinguishable as the number of layers increases.
To tackle the problem,
existing methods
utilize various techniques,
such as residual connection~\cite{kipf2016semi}, normalization schemes~\cite{zhao2019pairnorm} and boosting strategy~\cite{sun2020adagcn}.
There are also
methods~\cite{DropEdge,chen2018fastgcn} 
that adopt graph augmentation to remove a proportion of nodes or edges in the graph to alleviate the problem.
Recently,
\citet{klicpera2018predict} and \citet{liu2020towards} present that the coupling of feature propagation and transformation causes the problem 
and they decouple the two steps to 
mitigate the effect of over-smoothing.
Further,
GCNII~\cite{chen2020simple} proposes initial residual and identity mapping which can effectively boost the model performance.

Moreover, 
there are also many studies on designing GNNs for graphs with heterophily~\cite{fagcn2021, chien2021adaptive, twocoins,li2022finding,zhu2020beyond}.
For example, 
H\textsubscript{2}GCN~\cite{zhu2020beyond} presents three strategies to
improve the performance of GNNs under heterophily: ego
and neighbor embedding separation, higher-order neighborhood utilization and intermediate representation combination.
After that,
based on the generalized PageRank, GPR-GNN~\cite{chien2021adaptive} combines the low-pass and high-pass convolutional filters by
adaptively learning the signed weights of node embeddings in each propagation layer.
Further,
to address the heterophily issue,
\citet{li2022finding} propose to find global homophily for each node by taking all the nodes in the graph as the node's neighborhood.
Our proposed framework models the priority of a node as a function of the heterophily degree of the node's neighborhood,
which can improve the performance of GNNs on heterophilous graphs.

\subsection{Learning to Propagate in GNNs}
Since different nodes in a graph may need a personalized number of propagation layers,
learning to propagate in GNNs has recently attracted much attention.
Although it can be used to alleviate the effect of over-smoothing,
they are two different problems because learning to propagate targets at learning the propagation strategies of nodes.
Currently, some methods~\cite{xu2018representation, xiao2021learning, wang2022graph, sun2020adagcn} have been proposed to learn message propagation in GNNs.
For example, JKnet~\cite{xu2018representation} presents an architecture that flexibly leverages different neighborhood ranges for each node to enable better structure-aware representation.
Further, \citet{xiao2021learning} present a general learning framework which can explicitly learn the interpretable and personalized propagation strategies for different nodes. 
Recently, \citet{wang2022graph} present a framework that uses parametric controllers to decide the propagation depth for each node based on its local patterns.
Although these works 
aim to learn personalized propagation strategies for each node,
they ignore node priority that can be reflected by node influence and heterophily.
This further motivates our study on prioritized node-wise message propagation.

\section{Preliminary}
In this section, we
introduce the notations used in this paper and also some GNN basics.
\subsection{Notations}
Let $G = \mathcal{(V, E)}$ denote an undirected graph without self-loops, 
where $\mathcal{V} = \left \{  v_{i} \right \}_{i=1}^{n} $ is a set of nodes
and $\mathcal{E\subseteq V\times V }$ is a set of edges.
Let $A$ be the adjacency matrix of $G$ such that 
$A_{ij} =1$ if there exists an edge between nodes 
$v_{i}$ and $v_{j}$; 0, otherwise.
We denote the set of adjacent neighbors of node $v_i$ as 
$\mathcal{N}_{i}$.
We further define a diagonal matrix $D$, where $D_{ii}=\textstyle \sum_{j=1}^{n}A_{ij} $ is the degree of node $v_{i}$.
We use $X\in \mathbb{R}^{n\times d}$ to denote the node feature matrix,
where the $i$-th row $X_{i}$ is the $d$-dimensional feature vector of node $v_{i}$.
For the node representation matrix in the $k$-th layer,
we denote it as $ H^{(k)} $, where the $i$-th row $h_{i}^{(k)}$ is the corresponding embedding vector of node $ v_{i}$.
We also use $ Y \in \mathbb{R}^{n\times c}$ to denote the ground-truth node label matrix, where $c$ is the number of labels.
In this paper,
we focus on the node-level classification task, 
where each node $v_{i}$ is associated with a label {$y_{i}$}.

\subsection{Message Propagation}

GNNs aggregate messages from {a node's} neighborhood by different message propagation strategies. 
Generally, 
each propagation step in
GNNs
includes two sub-steps.
We take an arbitrary node $v_i$ in the $k$-th layer as an example.
The first sub-step is to aggregate information from a node's neighbors,
which is given as:
\begin{equation} 
\hat{h} _{i}^{(k)}=\texttt{AGGREGATE}(h_{j}^{(k-1)}, \forall v_{j}\in \mathcal{N}_{i}). 
\end{equation}
After that,
the second sub-step is to update node embeddings:
\begin{equation} 
h_{i}^{(k)}=\texttt{UPDATE}(h_{i}^{(k-1)}, \hat{h} _{i}^{(k)}). 
\end{equation}
Many state-of-the-art GNN models follow this message propagation mechanism, 
such as GCN~\cite{kipf2016semi} and GAT~\cite{velivckovic2017graph}. 
Further,
there are also some methods that add the initial node embedding $h_i^{(0)} =  f_{\theta } (X_i)$ in the $\texttt{UPDATE}$ function, 
such as APPNP~\cite{klicpera2018predict} and GCNII~\cite{chen2020simple}.
In this case,
the \texttt{UPDATE} function should be modified into $h_{i}^{(k)}=\texttt{UPDATE}(h_{i}^{(k-1)},\hat{h}_{i}^{(k)}, h_{i}^{(0)} )$.
After $L$ propagation steps, the final node embedding $h_i^{(L)}$ will be used in downstream tasks.

\begin{figure}[h]
  \centering  \includegraphics[width=\linewidth]{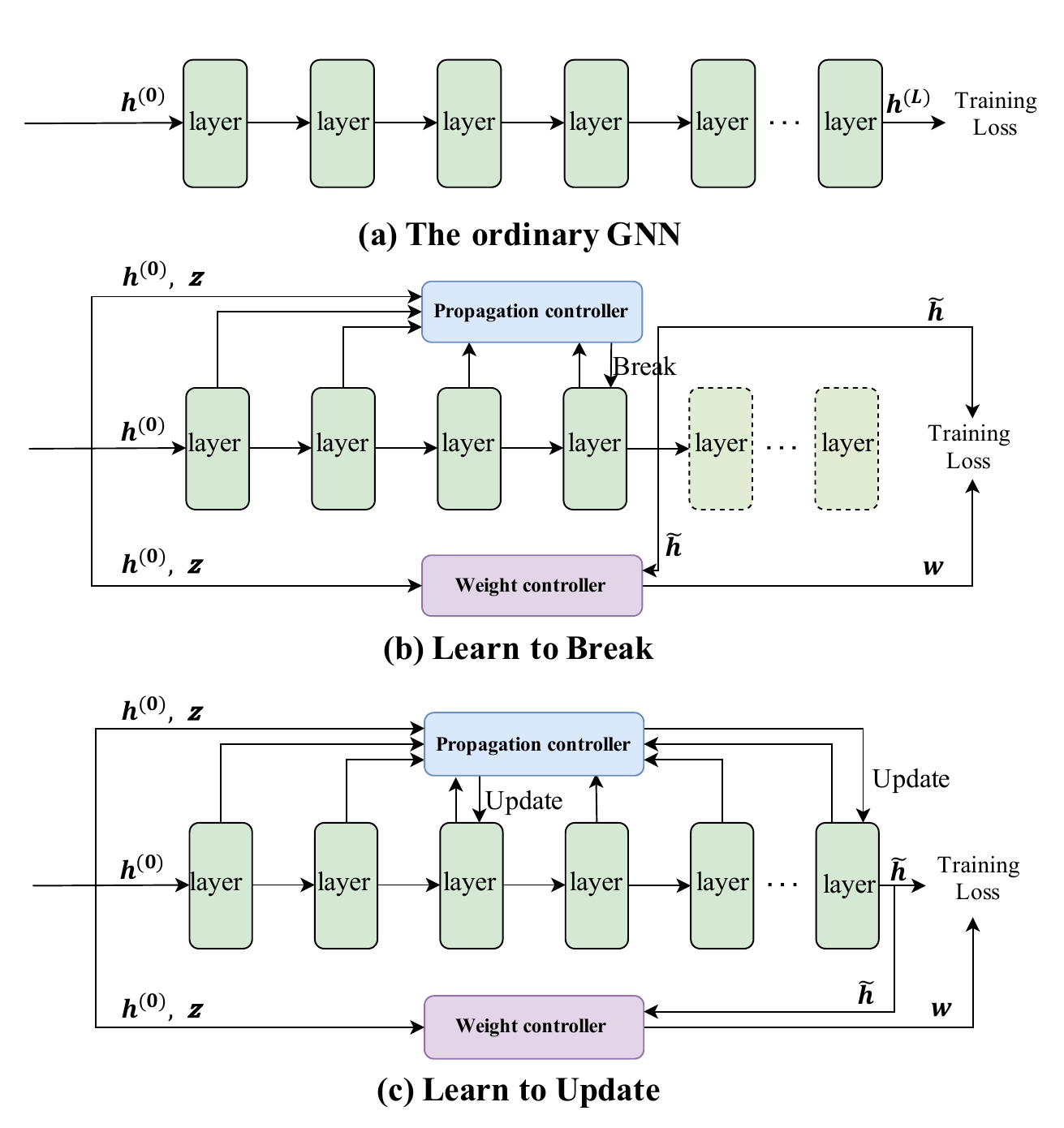}
  \caption{The overall framework of \name.
  The propagation controller learns the propagation steps for nodes while the 
  weight controller calculates the priority scores. (a) The general GNN pipeline; (b) Learning to break; (c) Learning to update}
  \label{fig:model}
\end{figure}

\section{Method}
In this section,
we propose our framework \name,
which consists of a backbone GNN model, a propagation controller 
to learn propagation depth for nodes
and a weight controller to learn node priority.
We first introduce priority measures for nodes 
and then describe prioritized message propagation learning in propagation controller.
After that,
we show how to learn priority weights for nodes in weight controller.
Finally,
we present the 
optimization algorithm.
The overall framework of \name\ is summarized in Figure~\ref{fig:model}.

\subsection{Priority Measures}
\label{sec:prio}
As we discussed in {Section~\ref{introduction}}, 
we would like to measure the priority of a node w.r.t. the degree of node influence and neighborhood heterogeneity.
Specifically, 
we introduce three priority measures, 
including \emph{degree centrality}, 
\emph{eigenvector centrality}, and 
\emph{heterophily degree}. 

\textbf{Degree Centrality}. 
A node's degree is the number of edges connected with it (in the undirected graphs).
It is 
a measure for node centrality which is simple but effective \cite{10.1093/acprof:oso/9780199206650.001.0001}.
For example, 
for celebrities in social networks,
they often connect with a significant number of neighbors
and have higher influence than normal users.
The degree centrality of node $v_{i}$ can be defined as:
\begin{equation} 
d_{i} = {\textstyle \sum_{j=1}^{n}A_{ij} }. 
\end{equation}

\textbf{Eigenvector Centrality}. 
The eigenvector centrality is an extension to the degree centrality \cite{10.1093/acprof:oso/9780199206650.001.0001}.
Specifically,
it measures the centrality of a node
by 
distinguishing the importance of the node's neighbors, while the degree centrality only considers equally important neighbors.
The eigenvector centrality of a node
is the weighted average of that of its neighbors,
{which is formally formulated as}:
\begin{equation} 
ec_{i} = {\textstyle \sum_{j=1}^{n}A_{ij}ec_{j} }. 
\end{equation}

\textbf{Heterophily Degree}. 
Given a node $v_i$ with heterophilous neighborhood,
it usually needs a large propagation step to collect useful information from distant neighbors.
Further, 
compared with nodes that have homophilous neighbors,
the label of $v_i$ may be harder to predict
and it should thus be given higher priority in the learning process. 
Therefore, we introduce heterophily degree as a measure of node priority,
which is calculated by
\begin{equation} 
he_{i} = \left \|h_{i}^{(0)} - \hat{h}_{i}^{(1)}    \right \|_{2},  
\end{equation}
where $\hat{h}_{i}^{(1)} = \texttt{AGGREGATE}(h_{j}^{(0)}, \forall v_{j}\in \mathcal{N}_{i}) $.
Finally, 
we concatenate these three measures 
which are further used 
to learn propagation steps and priority weights in GNNs:
\begin{equation} 
\label{eq:zi}
z_{i} = \left [ d_{i}\parallel ec_{i} \parallel he_{i}  \right ]. 
\end{equation}
where $\parallel $ is the concatenation operator.

\subsection{Propagation Controller}
As shown in Figure~\ref{fig:model}(a), 
given a node with feature vector $h^{(0)}$ and 
a pre-set maximum propagation step $L$,
GNNs aggregate messages from its neighbors 
by a propagation strategy 
and output the final representation $h^{(L)}$.
However, it is inappropriate to set the same propagation step for
all the nodes. 
Therefore, 
we propose a propagation controller 
to learn personalized propagation step for each node,
which is implemented by two methods, 
namely,
\emph{learning to break} (L2B)
and \emph{learning to update} (L2U).
In both methods,
we use
node priorities 
and the difference between a node's representation and its neighbors' in each layer
to calculate node propagation steps.
Next, we describe the details of
both L2B and L2U.

\textbf{L2B}: The L2B method directly learns whether to break message propagation in a layer.
Let $ \mathcal F_{p}(\cdot; {\phi_p})$ be the propagation controller function 
with learnable parameters $ \phi_{p}$.
We also use $ {r}_b^{(k)} \in \{0,1\}^{n} $ 
to denote whether to break aggregating for all the nodes in the $k$-th layer. 
We calculate the probability of node $v_i$
that breaks message aggregation in the $k$-th layer by :
\begin{equation} 
Pr(r_{bi}^{(k)}=1) =  \mathcal F_{p}(z_{i},h_{i}^{(0)},  h_{i}^{(k)}, \hat{h}_{i}^{(k)}; {\phi_p}),  
\end{equation}
where $z_i$ derived from Eq.~\ref{eq:zi} represents the priority of $v_i$,
$h_{i}^{(0)}$ is 
$v_i$'s initial embedding 
and $\hat{h}_{i}^{(k)} = \texttt{AGGREGATE}(h_{j}^{(k-1)}, 
\forall v_{j}\in \mathcal{N}_{i}) $ is the representation of $v_i$'s neighborhood in the $k$-th step.
After the probability is calculated,
we introduce
a hyper-parameter $\varepsilon$
as a threshold 
to control whether message propagation should break.
Specifically,
if $P({r}_{bi}^{(k)}=1 ) >\varepsilon$,
we break information aggregation for $v_i$ from the $k$-th layer and obtain its final embedding
$\tilde{h}_i = h^{(k)}_i$;
otherwise,
message propagation continues.

\textbf{L2U}: 
Different from L2B,
L2U does not stop message propagation until it traverses all the $L$ steps.
We overload $\tilde{h}_i$
as the best embedding of $v_i$ 
in all the traversed steps.
In the $k$-th step,
for node $v_i$,
L2U judges whether 
node embedding 
$h^{(k)}_i$ is better than $\tilde{h}_i$.
If yes,
we update $\tilde{h}_i = h^{(k)}_i$
and record the corresponding propagation step $k$ for node $v_i$;
otherwise,
we keep $\tilde{h}_i$ unchanged.
To characterize 
the advantage of $h^{(k)}_i$
over $\tilde{h}_i$,
we add $\tilde{h}_i$
as an additional input to the propagation controller function 
$ \mathcal F_{p}$.
Formally,
we use 
$ {r}_{ui}^{(k)} \in \{0,1\} $ to
represent whether to update $\tilde{h}_i$ 
in the $k$-th step.
Then the probability
to update $\tilde{h}_i$
is computed by:
\begin{equation} 
Pr({r}_{ui}^{(k)}=1) =  \mathcal F_{p}(z_{i}, h_{i}^{(0)}, h_{i}^{(k)}, \hat{h} _{i}^{(k)}, \tilde{h}_{i}; {\phi_{p}}). 
\end{equation}
After that,
if $Pr({r}_{ui}^{(k)}=1) > \varepsilon$,
we then update $\tilde{h}_i = h^{(k)}_i$.
The above process is repeated until $L$ message propagation steps are performed. 
Finally,
L2U sets the propagation step for node $v_i$
as 
the largest step where 
$\tilde{h}_i$ is updated.
Note that, in both L2B and L2U,
$\mathcal F_{p}$
can be implemented using MLP.

We design two different loss functions for L2B and L2U, respectively. 
For L2B, 
it learns whether to break message propagation in a given step.
Intuitively, 
for a node that has a small prediction error,
it is expected to stop 
message aggregation.
Therefore,
we normalize the prediction error and 
compare it with the probability of continuing message propagation.
Formally,
we define the loss for L2B as:
\begin{equation} 
\label{eq:lbp}
\mathcal L_{p}^b =\frac{1}{m}  {\textstyle \sum_{i=1}^{m}
\left | \frac{C_{i} -C_{min}}{C_{max}-C_{min}} -Pr(r_{bi}^{(l_{i})}=0)  \right | }, 
\end{equation}
where $C_{i} = C(y_{i},\hat{y}_{i})$ and $C_{max},C_{min}$ is the max value and min value of $C$. 
Note that 
$l_{i}$ is the step to break the propagation for node $v_i$ and
$r_{bi}^{(l_i)}=0$ means to continue aggregation for $v_{i}$ in the $l_i$-th step.
From Equation~\ref{eq:lbp},
when $C_i$ is small,
$Pr(r_{bi}^{(l_{i})}=0)$ is also small, i.e., message should stop propagating;
otherwise,
$Pr(r_{bi}^{(l_{i})}=0)$ is large,
which indicates that message should continue propagation.

For L2U, 
it learns the best embedding of a node in all the pre-set $L$ steps.
Intuitively,
for a node whose best embedding $\tilde{h}_i$ is to be updated with node embedding $h_{i}^{(k)}$ in the $k$-th step,
it indicates that 
$Pr(r_{ui}^{(k)}=1)$ is large
and 
the prediction error $C_{i}^{(k)}$ derived from $h_{i}^{(k)}$ should be smaller than the prediction error $\tilde{C}_{i}$ derived from $\tilde{h}_i$;
otherwise,
$Pr(r_{ui}^{(k)}=1)$ is small
and $C_{i}^{(k)}$ should be larger than $\tilde{C}_{i}$.
Hence,
the loss function for L2U can be  formulated as:
\begin{equation} 
\mathcal L_{p}^u =\frac{1}{m}  {\textstyle \sum_{i=1}^{m}\frac{1}{L} 
{\textstyle \sum_{k=1}^{L}Pr(r_{ui}^{(k)}=1)(C_{i}^{(k)} - \tilde{C}_{i})}}.
\end{equation}
For brevity,
we use $\mathcal L_{p}$ to denote the loss function of propagation controller for both L2B and L2U in the following.


\subsection{Weight Controller}
To assign different priorities to nodes in the graph,
we present an adaptive weight controller.
On the one hand,
the priority of a node is directly related to the measures given in Section~\ref{sec:prio}.
On the other hand,
with the learning process,
the propagation controller will output 
the corresponding propagation step for each node.
Intuitively,
the larger the propagation step,
the more difficult a node to be classified, 
and the larger priority should be given.
We also use 
node embeddings to help compute node priority. In particular, for node $v_i$,
we use
$h_i^{(0)}$ and $\tilde{h}_i$.
Specifically,
we define $ \mathcal F_{w}(\cdot;\phi_w)$ as the weighting function with learnable parameters $\phi_{w}$,
and
calculate the priority weight $w_{i}$ of a node $v_{i}$ by:
\begin{equation} 
\label{eq:fw}
w_{i} = \mathcal F_{w} (z_{i}, h_{i}^{(0)},\tilde{h}_{i},l_{i}; \phi_{w}), 
\end{equation}
where 
$l_{i}$ is the propagation step output by the propagation controller 
for node $v_{i}$. 
$\mathcal F_{w}$
can be implemented with MLP.
After $w_i$ is computed,
we inject it into the loss function used by GNN models.
Given a backbone GNN model
$ \mathcal F_{g}$ with
learnable parameters $\theta$,
the general loss function in the supervised learning is given as
\begin{equation} 
\mathcal L_{g} = \frac{1}{m}{\textstyle \sum_{i=1}^{m}C(y_{i}, \hat{y}_{i})}, 
\end{equation}
where $\hat{y}_{i}= \mathcal F_{g}(x_{i};\theta)$ is the predicted label for node $v_i$, $y_i$ is the true label of $v_i$, and $C(y_{i}, \hat{y}_{i})$ is a distance measure and $m$ is the training set size. 
With the weight $w_{i}$, 
we aim to minimize the reweighted loss function:
\begin{equation} 
\label{eq:obj}
\mathcal{L}_g = 
\max\limits_{w_i}
\min\limits_{\theta} \frac{1}{m}  {\textstyle \sum_{i=1}^{m}w_{i}C(y_{i}, \hat{y}_{i})  } - \lambda _{1} \frac{1}{m}  {\textstyle \sum_{i=1}^{m}w_{i}^{2}}. 
\end{equation}
Here,
the second term is $l_2$-norm and 
$\lambda_1$ is a hyper-parameter that controls the importance of the two terms.

\subsection{Optimization}
From the discussion above,
we see that 
our framework 
has three learnable parameters:
$\theta$, 
$\phi_p$ 
and $\phi_w$,
which are coupled with each other.
Both
the two controllers 
take node embeddings output by the backbone GNN model as input, 
while the backbone GNN model depends on the two controllers to calculate priority weight
and propagation step. 
Therefore,
we adopt an alternating optimization strategy that optimizes each parameter with others fixed.

First,
to optimize $\theta$,
we fix $w_i$ in Equation~\ref{eq:obj}
and 
directly calculate $\frac{\partial \mathcal{L}_g}{\partial \theta}$.
Then
for the weight controller,
we fix $\theta$
and reduce Eq.~\ref{eq:obj} 
to $\mathcal{L}_w = 
\max\limits_{w_i} \frac{1}{m}  {\textstyle \sum_{i=1}^{m}w_{i}C(y_{i}, \hat{y}_{i})  } - \lambda _{1} \frac{1}{m}  {\textstyle \sum_{i=1}^{m}w_{i}^{2}}
$.
We compute $\frac{\partial \mathcal{L}_w}{\partial {\phi_{w}}}$ and 
adopt stochastic gradient ascent to optimize $\phi_{w}$. 
In this way,
a large weight will be given to the node with large error prediction.
Further,
we fix $\theta$ and $\phi_{w}$,
and optimize $\phi_{p}$
in $\mathcal{L}_p$.

Since weight controller and propagation controller
share a large number of inputs
and they are optimized alternatively,
we implement them with a two-layer MLP in our experiments,
where they share parameters in the first layer 
and have their own parameters in the second layer.
We minimize the merged loss function of both controllers as:
\begin{equation} 
\mathcal L_{Controller} = \mathcal L_{p} - \lambda _{2} \mathcal L_{w},
\end{equation}
where $\lambda_2$ is used to balance the two terms.

\textbf{[Time complexity analysis].}
Our proposed framework \name\ 
can be integrated with any mainstream GNNs,
which additionally
introduces  
two controllers.
Since the two controllers are implemented with a two-layer MLP, 
the time complexity of \name\ is 
linear w.r.t. the number of nodes in the graph.
Finally,
we summarize the pseudocodes of our proposed framework PriPro in Algorithm~\ref{PriPros}
which can be found in Section~\ref{ap:a} of the appendix.

\section{EXPERIMENTS}
This section comprehensively 
evaluates the performance of \name\ against other state-of-the-art methods on benchmark datasets.
\subsection{Experimental Settings}
\textbf{Datasets.} 
We use 8 datasets in total,
which can be divided into two groups.
The first group is homophilous graphs,
which include
\emph{Cora}, \emph{Citeseer} and \emph{Pubmed}.
These datasets are three citation networks that are widely used for node classification~\cite{yang2016revisiting}.
In these datasets,  
nodes represent publications
and
edges are citations between them.
Further,
node features are the bag-of-words representations of keywords contained in the publications.
The other group of datasets is heterophilous graphs.
We adopt five public datasets
from~\cite{pei2020geom}: \emph{Actor}, 
\emph{Chameleon}, \emph{Cornell}, \emph{Texas}, \emph{Wisconsin}.
Specifically,
Actor is a graph with heterophily,
which represents actor co-occurrence in Wiki pages;
for other datasets,
they are web networks, 
where nodes are web pages and edges are hyperlinks.
Statistics of these datasets are summarized in Table~\ref{dataset} 
of Section~\ref{ap:b} in Appendix.

\noindent\textbf{Baselines.} 
To evaluate the effectiveness of 
\name, we compare it with the 
SOTA GNN models, 
including
GCN~\cite{kipf2016semi}, 
SGC~\cite{wu2019simplifying},  GAT~\cite{velivckovic2017graph}, 
JKnet~\cite{xu2018representation}, 
JKnet+ DropEdge~\cite{DropEdge} (JKnet (Drop)),
APPNP~\cite{klicpera2018predict}, 
GCNII~\cite{chen2020simple}, GCNII$^*$~\cite{chen2020simple}, and two representative heterophilous-graph-oriented methods:
H$_{2}$GCN~\cite{zhu2020beyond} and Geom-GCN~\cite{pei2020geom}. 
We further compare \name\ with
L2P~\cite{xiao2021learning}, which is the SOTA learn-to-propagate framework.
For NW-GNN~\cite{wang2022graph},
while it can also learn 
personalized propagation for nodes,
it focuses on the whole neural architecture search for various components in GNNs.
Since its codes are not publicly available,
we do not take it as our baseline for fairness.
For our framework \name,
since we have two different strategies in the propagation controller,
we name the corresponding frameworks as L2B and L2U for short, respectively.
For fairness, we use APPNP as our backbone, 
which is the same as L2P~\cite{xiao2021learning}, and our framework is also applicable to other GNN models~\cite{kipf2016semi,wu2019simplifying,velivckovic2017graph,chen2020simple}.


\subsection{Performance Comparison with GNNs}
We first conduct experiments to compare \name\ with other GNN models on both homophilous and heterophilous graphs.

\noindent\textbf{Performance on homophilous graphs.} 
We use the standard fixed training/validation/testing splits~\cite{kipf2016semi} on three datasets Cora, Citeseer, and Pubmed, with 20 nodes per class for training, 500 nodes for validation and 1,000 nodes for testing, respectively.
Table~\ref{homophily GNN} reports the mean classification accuracy on the test set after 10 runs.
From the table, we make the following observations:
(1) Our methods L2B and L2U perform very well on all the datasets.
In particular,
L2B achieves the best results on both Citeseer and Pubmed.
While it is not the winner on Cora,
the performance gap with the best result is marginal.
(2) 
Compared with the backbone model APPNP, 
our frameworks L2B and L2U can consistently improve its performance on all three datasets.
All these results show the effectiveness of \name\ on homophilous graphs.
\begin{table}[h]
  \caption{Node classification accuracy ($\%$) on homophilous graphs. We highlight the best score for each dataset in bold.}
  \label{homophily GNN}
  \begin{tabular}{c c c c}
    \toprule
    \textbf{Method} & \textbf{Cora} & \textbf{Citeseer} & \textbf{Pubmed} \\
    \midrule
    GCN & 81.5 & 70.3 & 79.0 \\
    SGC & 81.0 & 71.9 & 78.9 \\
    GAT & 83.0 & 72.5 & 79.0 \\
    JKNet & 81.1 & 69.8 & 78.1 \\
    JKNet (Drop) & 83.3 & 72.6 & 79.2 \\
    APPNP & 83.3 & 71.8 & 80.1 \\
    GCNII & \textbf{85.5} & 73.4 & 80.2 \\
    GCNII$^*$ & 85.3 & 73.2 & 80.3 \\
    \midrule
    L2B & $85.0\pm0.8$ & $\textbf{74.0}\pm0.8$ & $\textbf{81.1}\pm0.4$ \\
    L2U & $85.0\pm0.4$ & $72.7\pm0.9$ & $81.1\pm0.5$ \\
  \bottomrule
\end{tabular}
\end{table}

\noindent\textbf{Performance on heterophily graphs.} Table~\ref{heterophily GNN} shows the mean classification accuracy of all the models on heterophilous graphs. 
Here,
we use four datasets: Chameleon, Cornell, Texas and Wisconsin. 
For each dataset, 
we randomly split nodes of each class into 60\%, 20\%, and 20\% for training, validation and testing, respectively.
We measure the model performance on the test sets over 10 random splits as suggested in \cite{pei2020geom}. 
From the table, we 
see that (1) 
Our framework can consistently lead to the best results on all the datasets.
In particular,
L2B is the winner for three cases while L2U beats others on Wisconsin.
(2)
Our framework shows superiority over H$_2$GCN and Geom-GCN, which are specially designed for graphs with heterophily.
(3) 
Both L2B and L2U significantly improve APPNP 
on heterophilous graphs.
These results show that even in heterophilous graphs, 
our framework can still work well.
This is because 
we give high priorities to nodes with heterophilous neighborhood that are difficult to classify in the learning process.
\begin{table}[h]
  \caption{Node classification accuracy ($\%$) on heterophilous graphs. For each dataset, we highlight the best result in bold.}
  \label{heterophily GNN}
  \begin{tabular}{c c c c c}
    \toprule
    \textbf{Method} & \textbf{Chameleon} & \textbf{Cornell} & \textbf{Texas} & \textbf{Wisconsin} \\
    \midrule
    GCN & 28.18 & 52.70 & 52.16 & 45.88 \\
    GAT & 42.93 & 54.32 & 58.38 & 49.41 \\
    JKNet & 60.07 & 57.30 & 56.49 & 48.82 \\
    JKNet (Drop) & 62.08 & 61.08 & 57.30 & 50.59 \\
    APPNP & 54.30 & 73.51 & 65.41 & 69.02 \\
    GCNII & 60.61 & 74.86 & 69.46 & 74.12 \\
    GCNII$^*$ & 62.48 & 76.49 & 77.84 & 81.57 \\
    H$_{2}$GCN & 57.11 & 82.16 & 84.86 & 86.67 \\
    Geom-GCN & 60.90 & 60.81 & 67.57 & 64.12 \\
    \midrule
    L2B & $\textbf{63.68}$ & $\textbf{86.22}$ & $\textbf{85.67}$ &  $85.29$ \\
    L2U & $54.42$ & $84.32$ & $85.67$ & $\textbf{86.86}$ \\
  \bottomrule
\end{tabular}
\end{table}

\noindent\textbf{Performance with other backbones.} 
To further show the effectiveness of our framework, we use other GNN models as backbone, including GCN~\cite{kipf2016semi}, GAT~\cite{velivckovic2017graph} and GCNII~\cite{chen2020simple}.
Table~\ref{other backbone} reports the 
classification results.
From the table, we 
see that both L2B and L2U can consistently
improve the performance of GCN and GAT on all three datasets.
For GCNII,
while it is the state-of-the-art GNN model, L2B can also enhance its performance on Citeseer and Pubmed.
We further notice that
L2U is outperformed by GCNII.
This could be explained by the fact that 
GCNII adopts the {initial residual} and {identity mapping} techniques 
to preserve the initial features and information in previous layers for each node in each propagation layer.
This leads to a small difference between node embeddings in consecutive propagation layers, which increases the difficulty for L2U to predict the layer to be selected and adversely affect its performance.
In general, our framework can work well with other backbone models, which demonstrates its effectiveness.
\begin{table}[h]
  \caption{Node classification accuracy ($\%$) on homophilous graphs with other backbones.}
  \label{other backbone}
  \begin{tabular}{c c c c }
    \toprule
    \textbf{Method} & \textbf{Cora} & \textbf{Citeseer} & \textbf{Pubmed} \\
    \midrule
    GCN & 81.5 & 70.3 & 79.0 \\
    L2B & 82.6 & \textbf{73.3} & \textbf{79.5} \\
    L2U & \textbf{82.7} & 72.7 & 79.4 \\
    \midrule
    GAT & 83.0 & 72.5 & 79.0 \\
    L2B & \textbf{83.3} & 73.0 & 79.0 \\
    L2U & 83.2 & \textbf{73.6} & \textbf{79.7} \\
    \midrule
    GCNII & \textbf{85.5} & 73.4 & 80.2 \\
    L2B & 84.6 & \textbf{73.9} & \textbf{80.4} \\
    L2U & 84.2 & 72.6 & 79.8 \\
  \bottomrule
\end{tabular}
\end{table}

\subsection{Over-smoothing}
To evaluate the effectiveness of our proposed framework for alleviating the over-smoothing problem, 
we study how L2B and L2U perform as the number of layers increases compared to other state-of-the-art GNNs.
Table~\ref{deep} summarizes the classification results.
We vary the number of layers from $\left \{ 2, 4, 8, 16, 32, 64 \right \}$.
From the table, 
it is clear that, 
as the number of propagation steps increases,
the performance of GCN drops rapidly, while
both L2B and L2U can consistently perform well on all three datasets.
Although GCNII can also alleviate over-smoothing,
its performance in small propagation steps is poor.
For example,
with two propagation steps,
CGNII achieves an accuracy of $68.2$ only on Citeseer,
while that of L2B is $72.9$.
This shows that our proposed framework can lead to more stable model performance in various propagation steps.
\begin{table}[h]
  \small
  \caption{Node classification accuracy ($\%$) on homophilous graphs with different propagation steps. For each method,
  we highlight its best score among all the layers in bold.}
  \label{deep}
  \begin{tabular}{c c | c c c c c c }
    \toprule
    \multirow{2}{*}{\textbf{Datasets}} & \multirow{2}{*}{\textbf{Method}} & \multicolumn{6}{c}{\textbf{Propagation steps}} \\ 
    & & 2 & 4 & 8 & 16 & 32 & 64 \\
    \midrule
    \multirow{7}{*}{Cora} & GCN & \textbf{81.5} & 80.4 & 69.5 & 64.9 & 60.3 & 28.7 \\ 
    & JKNet & - & 80.2 & 80.7 & 80.2 & \textbf{81.1} & 71.5 \\
    & JKNet (Drop) & - & \textbf{83.3} & 82.6 & 83.0 & 82.5 & 83.2 \\
    & GCNII & 82.2 & 82.6 & 84.2 & 84.6 & 85.4 & \textbf{85.5} \\
    & GCNII$^{*}$ & 80.2 & 82.3 & 82.8 & 83.5 & 84.9 & \textbf{85.3} \\
    & L2B & 82.9 & 84.1 & \textbf{85.0} & 84.6 & 84.8 & 84.6 \\
    & L2U & 83.2 & 84.2 & 84.6 & 84.7 & \textbf{85.0} & 84.5 \\
    \midrule
    \multirow{7}{*}{Citeseer} & GCN & \textbf{70.3} & 67.6 & 30.2 & 18.3 & 25.0 & 20.0 \\ 
    & JKNet & - & 68.7 & 67.7 & \textbf{69.8} & 68.2 & 63.4 \\
    & JKNet (Drop) & - & 72.6 & 71.8 & \textbf{72.6} & 70.8 & 72.2 \\
    & GCNII & 68.2 & 68.9 & 70.6 & 72.9 & 73.4 & \textbf{73.4} \\
    & GCNII$^{*}$ & 66.1 & 67.9 & 70.6 & 72.0 & \textbf{73.2} & 73.1 \\
    & L2B & 72.9 & 73.4 & 73.4 & \textbf{74.0} & 73.4 & 72.0 \\
    & L2U & 71.5 & 71.5 & 72.3 & 72.6 & \textbf{72.7} & 72.4 \\
    \midrule
    \multirow{7}{*}{Pubmed} & GCN & \textbf{79.0} & 76.5 & 61.2 & 40.9 & 22.4 & 35.3 \\ 
    & JKNet & - & 78.0 & \textbf{78.1} & 72.6 & 72.4 & 74.5 \\
    & JKNet (Drop) & - & 78.7 & 78.7 & 79.1 & \textbf{79.2} & 78.9 \\
    & GCNII & 78.2 & 78.8 & 79.3 & \textbf{80.2} & 79.8 & 79.7 \\
    & GCNII$^{*}$ & 77.7 & 78.2 & 78.8 & \textbf{80.3} & 79.8 & 80.1 \\
    & L2B & 79.9 & 80.4 & 80.7 & \textbf{81.1} & 80.6 & 80.9 \\
    & L2U & 79.5 & 80.3 & 80.7 & \textbf{81.1} & 80.7 & 80.1 \\
  \bottomrule
\end{tabular}
\end{table}

\begin{figure}[!htb]
  \centering  \includegraphics[width=1.0\linewidth]{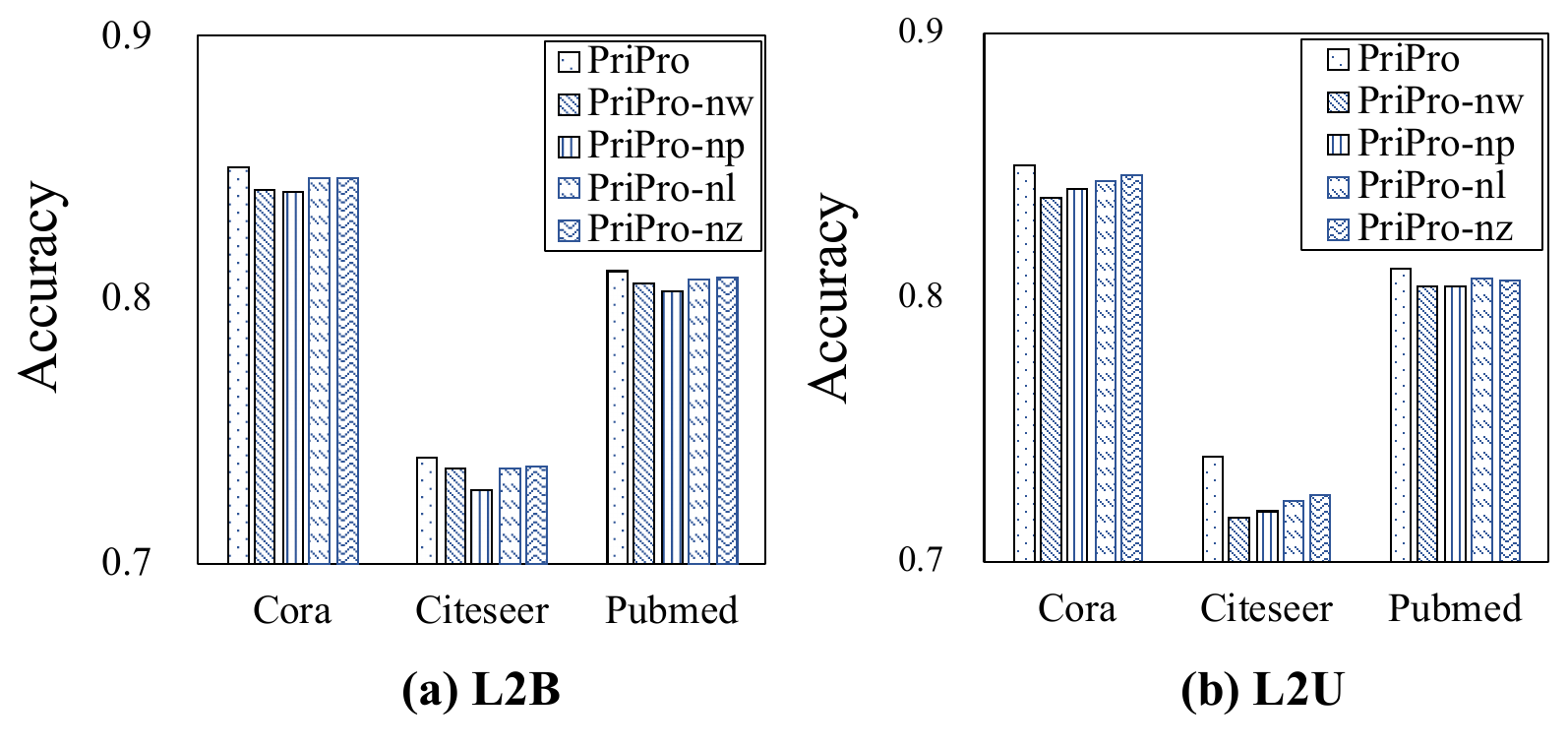}
  \caption{Ablation study}
  \label{Ablation}
\end{figure}
\begin{figure}[h]
  \centering  \includegraphics[width=0.9\linewidth]{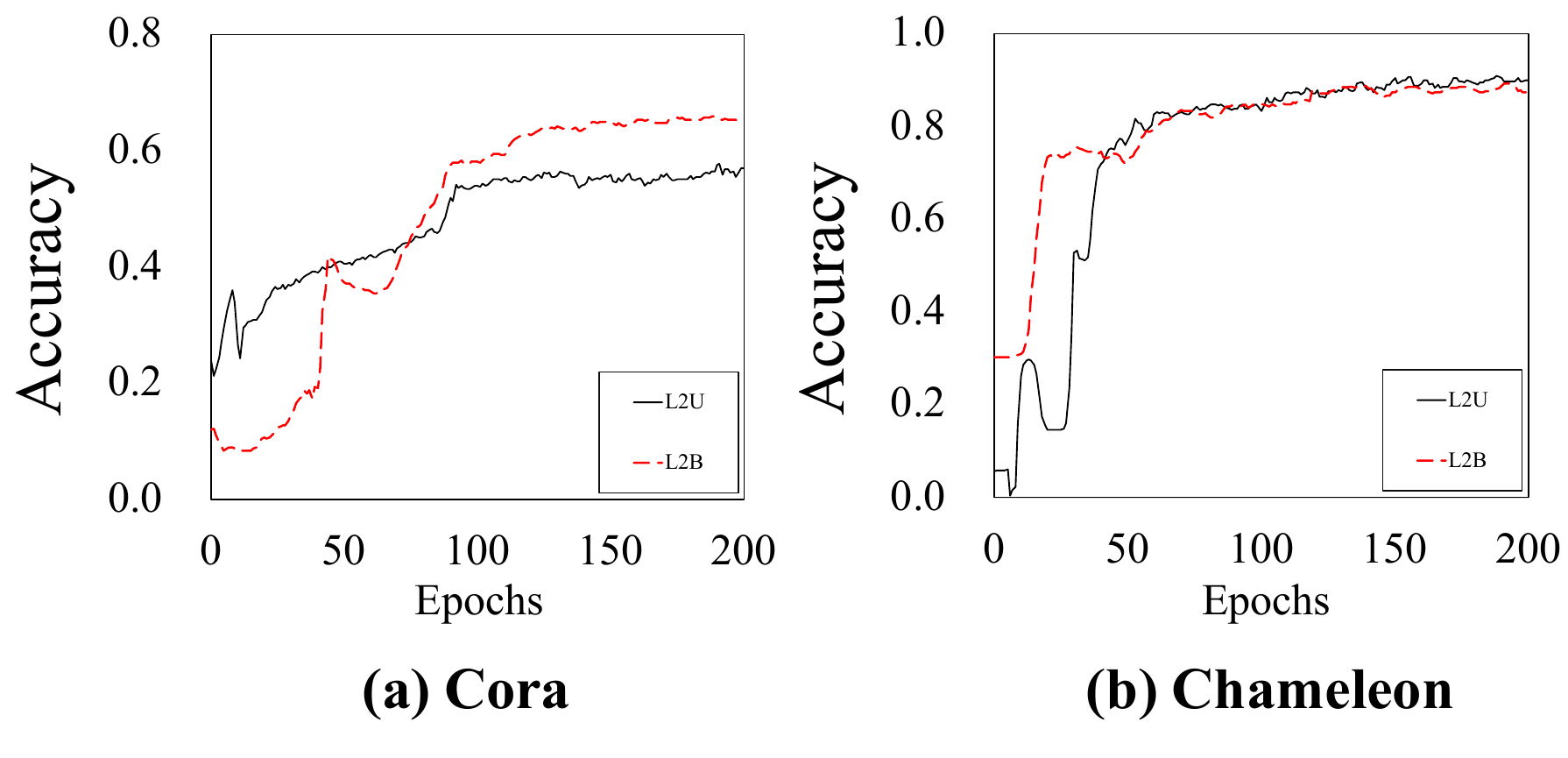}
  \caption{Convergence analysis}
  \label{Convergence}
\end{figure}
\subsection{Performance Comparison with Other Framework}
We further compare the performance of \name\ with L2P, 
which is the state-of-the-art framework for learning node-wise message propagation.
Specifically,
it uses two different strategies, namely, learning to quit (L2Q) and learning to select (L2S) to learn optimal propagation steps for nodes,
We first recap the difference between our proposed framework \name\ and L2P:
for L2B and L2Q,
both of them aim to learn whether to stop in a specific step,
but L2B considers node priority;
both L2U and L2S
learn in which step the node embedding should be selected as the final node representation
in a maximum of $L$ steps.
However,
L2S formulates the problem as a multi-class classification one,
which directly outputs the label (step) from $L$ class labels (steps),
while
L2U performs binary classification 
in each step.
For fairness, 
we 
take APPNP as the backbone model for both frameworks,
and directly report the results of L2S and L2Q from the original paper. 
Table~\ref{ta:l2p} 
illustrates the classification results.
From the table, 
we see that 
the \name\ framework leads to the best results in 5 out of 7 datasets,
which verifies the importance of incorporating node priority in learning node-wise message propagation.
In particular,
\name\ outperforms L2P on all heterophilous graphs,
which shows the effectiveness of 
giving high priorities to nodes with high heterophily.
Further,
L2U generally performs better than L2S.
This could be because 
when $L$ is large and labeled data is scarce,
it is difficult
to directly perform
$L$-class classification.


\begin{table*}[h]
\centering
\caption{Performance Comparison with L2P. For each dataset, we highlight the best result in bold.}
\small
\resizebox{0.7\linewidth}{!}
{
  \begin{tabular}{c c | c c c c c c c }
    \toprule
    \multirow{2}{*}{\textbf{Framework}} & \multirow{2}{*}{\textbf{Method}} & \multicolumn{7}{c}{\textbf{ Datasets}} \\ 
    & & Cora & Citeseer & Pubmed & Actor & Cornell & Texas & Wisconsin \\
    \midrule
    \multirow{2}{*}{L2P} 
    & L2Q & \textbf{85.2} & \textbf{74.6} & 80.4 & 37.0 & 81.1 & 84.6 & 84.7 \\
    & L2S & 84.9 & 74.2 & 80.2 & 36.6 & 80.5 & 84.1 & 84.3 \\ 
    \midrule
    \multirow{2}{*}{\name} & L2B & 85.0 & 74.0 & \textbf{81.1} & 36.8 & \textbf{86.2} & \textbf{85.7} & 85.3 \\
    & L2U & 85.0 & 72.7 & 81.1 & \textbf{37.1} & 84.6 & 85.7 & \textbf{86.9} \\ 
    
  \bottomrule
\end{tabular}
}
\label{ta:l2p}
\end{table*}

    
\begin{figure*}[h]
  \centering  \includegraphics[width=0.75\linewidth]{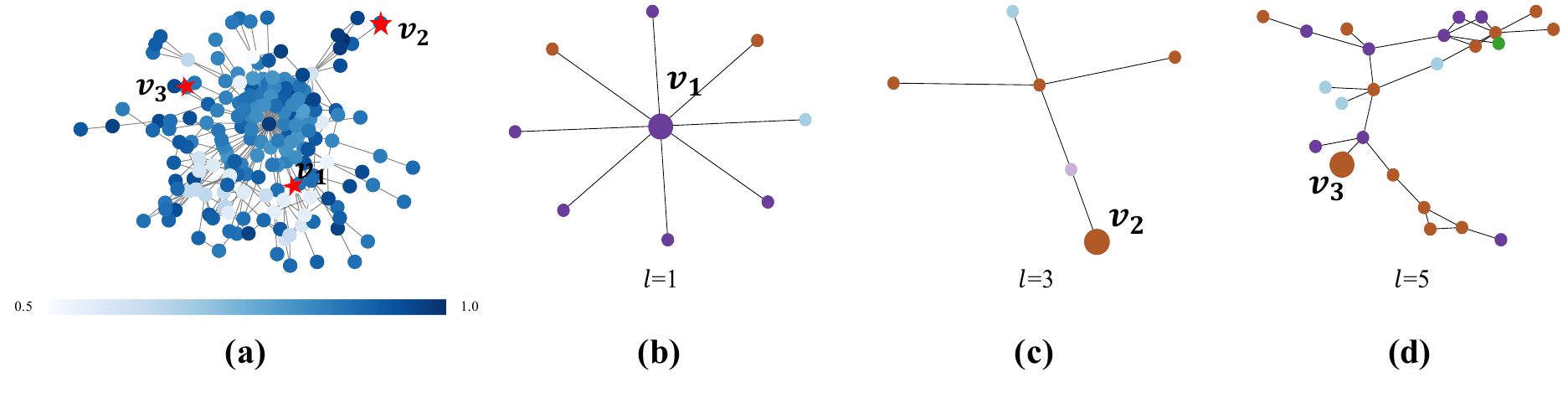}
  \caption{The visualization on the node priority weights and propagation steps calculated by L2B. In (b)-(d), we show the subgraphs centered at $v_1$, $v_2$ and $v_3$, respectively. 
  Different colors represent node labels.
  Better view in color.}
  \label{weights}
\end{figure*}
\begin{figure*}[h]
  \centering  \includegraphics[width=0.9\linewidth]{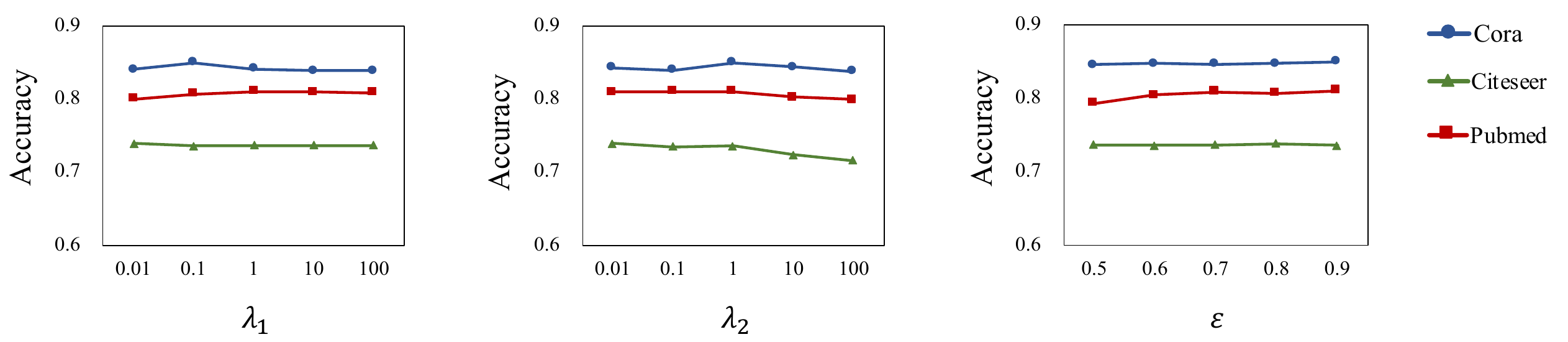}
  \caption{Hyper-parameter sensitivity analysis}
  \label{params}
\end{figure*}

\subsection{Ablation Study}
We next conduct an ablation study to understand the main components of PriPro,
which consists of a backbone GNN model,
a propagation controller and a weight controller.
Specifically, 
the propagation controller computes the optimal propagation step for each node considering node priority,
while 
the weight controller calculates the node priority scores by taking into account the optimal propagation steps of nodes. 
We first remove the optimal propagation step as input from the weight controller and call this variant \textbf{\name-nl} (\textbf{n}o propagation step \bm{$l$}).
Similarly, we remove the node priority as input from the propagation controller and call this variant \textbf{\name-nz} (\textbf{n}o node priority \bm{$z$}).
Further,
we remove the weight controller and 
the propagation controller,
and
call these variants \textbf{\name-nw} (\textbf{n}o \textbf{w}eight controller) and 
\textbf{\name-np} (\textbf{n}o \textbf{p}ropagation controller),
respectively.
Finally, we compare L2B and L2U with these variants 
and show the results in Figure~\ref{Ablation}.
From the figure, we see that
(1) \name\ clearly outperforms all the variants on the three datasets. 
(2) The performance gap between \name\ and 
\name-nw (\name-np) shows the importance of the weight controller (propagation controller) in learning node-wise message propagation. 
(3) \name\ performs better than \name-nl (\name-nz),
which shows that the optimal node propagation steps (node priority) 
can help learn better node priority (optimal node propagation steps) and further boost the model performance.

\subsection{Convergence Analysis}
We next analyze the convergence of \name\ during the learning process.
We conduct experiments on two datasets: Cora and Chameleon.
For other datasets,
we observe similar results, which are omitted
due to the space limitation.
We show the accuracy scores of both L2B and L2U on the validation set with increasing epochs
in Figure~\ref{Convergence}.
For both methods, 
we use the same training/validation set on each dataset and run the experiments for 200 epochs.
From the figure, we see that both L2B and L2U converge fast.
We speculate that this is because our proposed framework
implements both controllers with a two-layer MLP. 
The simple network structure of MLP mitigates the difficulty of model training, which further facilitates model convergence.

\subsection{Node Priority and Propagation Step}
To evaluate if \name\ can well learn node-wise priority and propagation steps, 
we visualize the results given by the weight controller and the propagation controller,
respectively.
Figure~\ref{weights}(a) shows the priority scores of nodes calculated by the weight controller of L2B.
For ornamental purposes, 
we take Cornell as an example, which has a small number of nodes and is easy to visualize. 
For all the nodes,
the darker the color, 
the larger the weight. 
From the figure, 
we find that the high-degree nodes tend to have large priority scores.
This is consistent with our
assignment of large priority for nodes with high influence.
Further,
we see that 
some leaf nodes with few links in the graph are also given large weights.
Since Cornell is a heterophilous graph,
these leaf nodes are often difficult to gather useful information 
from neighborhood.
This is also consistent with that we would like to focus more on nodes that are difficult to predict.
Figure~\ref{weights}(b)-(d)
further shows some case studies from Cornell on the learned node propagation steps,
where different colors represent node labels.
In Figure~\ref{weights}(b),
node $v_1$ has homophilous neighborhood where most adjacent neighbors are in the same label,
so 1-hop message aggregation is sufficient for label prediction.
Compared with node $v_1$,
node $v_2$ in Figure~\ref{weights} needs a larger propagation step due to its higher heterophily.
For node $v_3$ in Figure~\ref{weights}(d),
since its neighborhood is of large heterophily,
it needs a large propagation step $l=5$ to absorb useful information from distant nodes in the same label.
All these results demonstrate the effectiveness of the weight controller and the propagation controller in \name.

\subsection{Hyperparameter Sensitivity Analysis}
We end this section with a sensitivity analysis on the hyperparameters of \name.
In particular, we study three key hyperparameters: the regularization weight $ \lambda_1$,
the balance coefficient $\lambda_2$, and a threshold $\varepsilon$.
In our experiments, we vary one parameter each time with others fixed. 
Figure~\ref{params} illustrates the results of L2B on Cora, Citeseer and Pubmed w.r.t. classification accuracy.
From the figure, we see that 
for all three hyper-parameters,
L2B can 
give very 
stable performances over a wide range of hyperparameter values. 
The threshold $\epsilon$ determines the likelihood to break the propagation process. 
The larger the threshold, the easier to propagate more layers. 
Since \name\ can effectively alleviate the over-smoothing problem, increasing the threshold allows more steps to propagate, which would not affect the classification results too much.
All the results show the insensitivity of \name\ w.r.t. these hyper-parameters.

\section{CONCLUSIONS}
In this paper, we learned prioritized node-wise message propagation in GNNs and proposed the framework PriPro. 
The framework consists of three components: a backbone GNN model, a propagation controller to determine the optimal propagation steps for nodes, and a weight controller to compute the priority scores for nodes. 
We designed a mutually enhanced mechanism to compute node priority, optimal propagation step and label prediction. 
We conducted extensive experiments and compared PriPro with 11 other methods. 
Our analysis shows that PriPro is very effective and can lead to superior classification performance.

\bibliographystyle{ACM-Reference-Format}
\bibliography{sample-base}

\appendix

\newpage

\appendix
\section{ALGORITHM}
We summarize the pseudocodes of our proposed framework \name\ in Algorithm~\ref{PriPros}.
\label{ap:a}
\begin{algorithm}[h]
\caption{\name}
\label{PriPros}
    \LinesNumbered 
    \KwIn{
    Graph $G = \mathcal{(V, E)}$, learning rate $\alpha_{g}, \alpha_{\mathcal F}$, epoch $T$, max aggregation step $L$.}
    \KwOut{learned prameters $\theta, \phi$.}
    Randomly initialize $\theta^{(0)}, \phi^{(0)}$\;
    \For{$t=0,1,...,T-1$}{
        $\tilde{h}=h^{(0)}$\;
        \For{$k=1,2,...,L$}{
            $h^{(k)} =\texttt{GNN}(h^{(k-1)}, \theta^{(t)})$ \\
            $ Pr(r^{(k)}=1) = \mathcal F_{p}(\cdot;\phi_p^{(t)})$ \\
            $ \tilde{h} \gets Pr(r^{(k)}=1), h^{(k)}$ \\
        }
        $w = \mathcal F_{w}(\cdot;\phi_w^{(t)})$ \\
        $\theta^{(t+1)} = \theta^{(t)} - \alpha_{g}\nabla_{\theta}\mathcal L_{g}  $ \\
        repeat 3-9 with the parameter $\theta^{(t+1)}$ \\
        $\phi^{(t+1)} = \phi^{(t)}-\alpha_{\mathcal F}\nabla_{\phi} \mathcal L_{Controller} $ \\
    }
    \Return $\phi$ and $\theta$ 
\end{algorithm}

\section{DATASETS}
\label{ap:b}
In our experiments, we use the following real-world datasets,
whose details are given as follows.
Statistics of these datasets
are summarized in Table~\ref{dataset}.

\noindent\textbf{\emph{Cora}, \emph{Citeseer} and \emph{Pubmed}} are three homophilous graphs which are broadly applied as benchmarks.
In these datasets, each node represents a scientific paper and each edges represents a citation. These graphs use bag-of-words representations as the feature vector for each node. Each node is assigned a label indicating the research field.
The task on these datasets is node classification.

\noindent\textbf{\emph{Texas}, \emph{Wisconsin} and \emph{Cornell}} are three heterophilous graphs which representing links between web pages of the corresponding universities.
In these datasets, each node represents a web page and each edge represents a hyperlink between nodes. We take bag-of-words representations as a feature vector for each node.
The task on these datasets is node classification.

\noindent\textbf{\emph{Chameleon}} is a subgraph of web pages in Wikipedia. The task is to classify the nodes into five categories. Note that this dataset is heterophilous graphs.

\noindent\textbf{\emph{Actor}} is a heterophilous graph which represents actor co-occurrence in Wiki pages.
Node features are built from the keywords that are included in the actor's Wikipedia page.
Our task was to divide the actors into five classes.
\begin{table}[!htbp]
  \caption{Datasets statistics.}
  \label{dataset}
  \begin{tabular}{c c c c c}
    \toprule
    \textbf{Dataset} & \textbf{Classes} & \textbf{Nodes} & \textbf{Edges} & \textbf{Features}\\
    \midrule
    Cora & 7 & 2,708 & 5,429 & 1,433 \\
    Citeseer & 6 & 3,327 & 4,732 & 3,703 \\
    Pubmed & 3 & 19,717 & 44,338 & 500 \\
    Actor & 5 & 7,600 & 26,659 & 932 \\
    Chameleon & 5 & 2,277 & 36,101 & 2,325 \\
    Cornell & 5 & 183 & 295 & 1,703 \\
    Texas & 5 & 183 & 309 & 1,703 \\
    Wisconsin & 5 & 251 & 499 & 1,703 \\
  \bottomrule
\end{tabular}
\end{table}

\section{Implementation Details}
\label{ap:c}
We implement \name\ by PyTorch and optimize the framework by Adam~\cite{kingma2014adam}. 
For fairness, we use APPNP as our backbone, 
which is the same as L2P~\cite{xiao2021learning}, and our framework is also applicable to other GNN models~\cite{kipf2016semi,wu2019simplifying,velivckovic2017graph,chen2020simple}. 
We perform a grid search to fine-tune hyper-parameters 
based on the results on the validation set. 
Details on the search space can be found in Table~\ref{hyperparameters}.
Further, since the results of most baseline methods on these benchmark datasets are public, we directly report these results from their original papers. 
For those cases where the results are missing, 
we report their results from \cite{chen2020simple}.
We run all the experiments on a server with 32G memory and a single Tesla V100 GPU. 

\begin{table}[H]
  \caption{Grid search space.}
  \label{hyperparameters}
  \begin{tabular}{c | c }
    \toprule
    \textbf{Hyperparameter} & \textbf{Search space} \\
    \midrule
    lr & $ \{ 0.005, 0.01, 0.05, 0.1  \} $ \\
    dropout & $ [ 0, 0.9  ] $ \\
    early stopping & $ \{ 100, 200  \}$ \\
    weight decay &  \{1e-5, 5e-4, 1e-4, 5e-3, 1e-3   \}  \\
    $\lambda _1$ & $ \{ 0.01, 0.1, 1, 10^1, 10^2  \} $ \\
    $\lambda _2$ & $  \{ 0.01, 0.1, 1, 10^1, 10^2  \} $ \\
    $\varepsilon$ & $ [ 0.1, 0.9] $ \\
  \bottomrule
\end{tabular}
\end{table}

\end{document}